\documentclass[3p,times]{elsarticle}
\usepackage{multicol}
\usepackage{multirow}
\usepackage{booktabs}
\usepackage{textcomp}
\usepackage{caption}
\usepackage{graphicx}
\usepackage{algorithm}
\usepackage{subfigure}
\usepackage{textcomp}
\usepackage{amsmath,amssymb,amsfonts}

\begin{document}
\begin{frontmatter}
\title{Spatial-temporal associations representation and application for process monitoring using graph convolution neural network}
\author{Hao Ren$^{a}$}
\author{Xiaojun Liang$^{a, \ast}$}
\author{Chunhua Yang$^{b, a}$}
\author{Zhiwen Chen$^{b, a}$}
\author{Weihua Gui$^{b, a}$}

\cortext[mycorrespondingauthor]{Corresponding author: Xiaojun Liang; E-mail address: liangxj@pcl.ac.cn (X. J. Liang).}
\address[mymainaddress]{Department of Strategic and Advanced Interdisciplinary Research, Peng Cheng Laboratory, Shenzhen, Guangdong, 518055, China.}
\address[mysecondaryaddress]{the School of Automation, Central South University, Changsha, Hunan, 410083, China.}

\begin{abstract}
Thank you very much for the attention and concern of colleagues and scholars in this work. With the comments and guidance of experts, editors, and reviewers, this work has been accepted for publishing in the journal "Process Safety and Environmental Protection". The theme of this paper relies on the Spatial-temporal associations of numerous variables in the same industrial processes, which refers to numerous variables obtained in dynamic industrial processes with Spatial-temporal correlation characteristics, i.e., these variables are not only highly correlated in time but also interrelated in space. To handle this problem, three key issues need to be well addressed: variable characteristics modeling and representation, graph network construction (temporal information), and graph characteristics perception. The first issue is implemented by assuming the data follows one improved Gaussian distribution, while the graph network can be defined by the monitoring variables and their edges which are calculated by their characteristics in time. Finally, these networks corresponding to process states at different times are fed into a graph convolutional neural network to implement graph classification to achieve process monitoring. A benchmark experiment (Tennessee Eastman chemical process) and one application study (cobalt purification from zinc solution) are employed to demonstrate the feasibility and applicability of this paper.
\end{abstract}


\end{frontmatter}

\begin{multicols}{2}
\section{Background}
The problem and motivation have been illustrated in the abstract section of this paper, i.e., modern industrial processes generate many dynamic, associated, and multi-scale variables, which are more likely to implicit spatial-temporal associations knowledge for describing irregular changes at different times. This spatial-temporal association representation belongs to the implicit data knowledge in industrial processes.

Generally, industrial operation data has dynamism and correlation characteristics, which reflect the dynamic changes in working conditions, operations, and raw materials. When some abnormalities occur, the data distribution of the same variable varies at different periods usually presenting irregular changes on the one hand, and the relationships between different variables are also changing irregularly multiple variables in the same industrial process (implicit spatial-temporal correlations) on the other hand. This spatial-temporal correlation refers to multivariate variables obtained in dynamic industrial processes with spatial-temporal correlation characteristics, i.e., their variables are not only highly correlated in time but also interrelated in space [1].

Compared with general data knowledge, data knowledge in industrial processes is often related to mechanisms and processes knowledge which have much more richer process semantics. In particular, a simple anomaly or mistake may lead to serious consequences, which require a high level of requirement on higher reliability of data knowledge resulting in the extreme difficulty of data knowledge mining. Furthermore, this problem has attracted numerous scholars attentions around the world, such as the dynamic nature of monitoring variables are considered in the different data distribution for the same variable at different time periods to show irregular correlation changes [2-3].

Inspired, this paper utilize the spatial-temporal associations representation to achieve process monitoring. In detail, a static graph network snapshot at different time is constructed by numerous variables and their associations, and it can be fed into the graph convolutional neural network to achieve graph classification to achieve process monitoring. Using this method, changes (or abnormalities, or faults) in networks can be defined by the associations between different variables with irregular changes at different times. Finally, the feasibility and applicability are demonstrated by a benchmark and a practical removal process.

\section{Case Studies}
The theoretical section is no further elaboration here, interested parties can refer to the paper in journal "Process Safety and Environmental Protection". The case studies mainly refer to two issues: one benchmark and one practical application. The performance of these two studies demonstrates the feasibility and applicability of this proposed method.

\subsection{Benchmark experiment}
The benchmark of the chemical process of this paper is the Tennessee Eastman (TE) chemical process which had been widely employed to demonstrate the feasibility and effectiveness of some newly proposed methods in related literature. This process involves 52 monitoring variables (including 22 process measurement variables ($v_1$-$v_{22}$), 12 control parameters ($v_{23}$-$v_{33}$), and 19 indicator variables ($v_{34}$-$v_{52}$)) [4-6]. All of these variables owns dynamism and correlation characteristics which are suitable for the demonstration of this proposed method. Twenty-one faults and one normal status has been simulated to verify the feasibility and effectiveness of fault detection and diagnosis on process monitoring. The dataset version of this benchmark of the chemical process can be referred to in literature [4], where each state collects training and testing datasets. For the training dataset, 500 samples are recorded for the normal state while 480 samples for each state of 21 faults. For the testing dataset, 960 samples are collected for each state of 21 faults, and each fault occurs in the 160th sampling instant in each state. More details of process description have been elaborated in literature [5] and literature [6].

During the formal case study experiment, a series of hyper-parameters should be determined firstly, especially the sample number $L$, sliding window $\tau$, threshold hyper-parameter $T_{common}$, etc. It should be noted that numerous small tests are carried out to find the suitable hyper-parameters, such as the sample number $L$ is configured as $(15min) \times 4 / (3min)$ and it is set by numerous tests in figure 6. Related results and discussions had been elaborated in the journal paper. Finally, the sample number is configured as $L=20$, and the hyper-parameter threshold $T_{common}=0.6$.

Process monitoring in this benchmark refers to classification tasks to illustrate the effectiveness of this proposed method. This paper presents the idea of fault diagnosis by considering the spatial-temporal associations' knowledge which mainly focuses on perceiving the irregular dynamic changes for different variables at different times [7]. For the convenience of comparison and discussion, this paper employs four case studies to illustrate the outstanding performance. All the results of these four cases can be detailed as follows.

\begin{table}[H]
\centering
\footnotesize
\captionsetup{font={footnotesize}}
\caption{Fault diagnosis performance ($Acc$) of Case-1 (\%).}
\begin{tabular}{ccccc}
\toprule
Method           & $F^4$           &  $F^9$           & $F^{11}$          & Ave.              \\
\hline
GCN              & 61.7            & 57.2             &  51.3             & 56.7                 \\
FDA              & 75.5            & 8.5              &  53.0             & 45.7                 \\
LFDA             & 57.0            & 25.5             &  48.0             & 43.5                 \\
SLFDA            & 94.0            & 8.5              &  40.5             & 47.7                 \\
SKLFDA           & 81.0            & \textbf{100.0}   &  81.0             & 87.3                 \\
This method      & \textbf{97.4}   & 97.9             &  \textbf{94.4}    & \textbf{97.8}        \\
\bottomrule
\end{tabular}
\label{tab1}
\end{table}

\begin{table}[H]
\centering
\scriptsize
\captionsetup{font={footnotesize}}
\caption{Fault diagnosis performance ($Acc$) of Case-2 (\%).}
\begin{tabular}{ccccccc}
\toprule
method       & nor. & $F^2$ &  $F^6$ & $F^7$ & $F^{14}$ & Ave.  \\
\hline
GCN          & 70.7            & 70.7             &  81.7             & 53.8             & 79.6           & 71.3          \\
ICA+NB       & 95.5            & 95.0             &  75.0             & \textbf{100.0}   & 92.5           & 91.6          \\
DICA+NB      & 97.0            & 98.5             &  \textbf{100.0}   & 91.5             & 89.5           & 95.3          \\
ICA+CDMWNB   & \textbf{99.5}   & 94.5             &  \textbf{100.0}   & \textbf{100.0}   & 95.5           & 97.9          \\
DICA+CDMWNB  & 97.5            & 98.0             &  \textbf{100.0}   & \textbf{100.0}   & 99.5           & 99.0          \\
This method   & 98.1           & \textbf{100.0}   &  \textbf{100.0}   & 99.7             & \textbf{99.7} & \textbf{99.2} \\
\bottomrule
\end{tabular}
\label{tab2}
\end{table}

\begin{table}[H]
\centering
\footnotesize
\captionsetup{font={footnotesize}}
\caption{Fault diagnosis performance ($Acc$) of Case-3 (\%).}
\begin{tabular}{cccccc}
\toprule
Fault       & GCN  & SOM   &  CCA-SOM         &  B-LSTM &  This method      \\
\hline
normal      & 38.2 &60.0  & \textbf{100.0}   &  54.9   &  94.8            \\
$F^1$       & 55.7 &92.5  &   97.5           &  97.8   &  \textbf{98.5}   \\
$F^2$       & 66.4 &97.5  &   97.5           &  97.3   &  \textbf{97.9}   \\
$F^4$       & 42.9 &65.0  &   97.5           &  57.9   &  \textbf{98.4}   \\
$F^5$       & 42.3 &60.0  &   97.5           &  82.9   &  94.3            \\
$F^6$       & 82.2 &97.5  & \textbf{100.0}   &  97.6   &  97.8            \\
$F^7$       & 25.5 &42.5  &   95.0           &  96.6   &  \textbf{98.6}   \\
Ave.     & 50.5 &73.6  & \textbf{97.9}    &  83.6   &  \textbf{97.5}   \\
\bottomrule
\end{tabular}
\label{tab3}
\end{table}

\begin{table}[H]
\centering
\footnotesize
\captionsetup{font={footnotesize}}
\caption{Fault diagnosis performance ($Acc$) of Case-4 (\%).}
\begin{tabular}{cccccc}
\toprule
Fault       & GCN  &FDA   &   MRDA           &  dyn-MRDA      & This method      \\
\hline
$F^3$       & 54.4 &54.0  &   97.1           &  93.6              & \textbf{98.3}   \\
$F^4$       & 57.0 &86.9  &   98.6           &  98.6              & 95.8            \\
$F^{11}$    & 55.3 &41.1  &   87.2           &  96.0              & \textbf{96.8}   \\
Average     & 55.5 &60.7  &   94.3           &  96.1              & \textbf{96.8}   \\
\hline
$F^1$       & 67.7 &97.0  &   93.8           &  93.0              & \textbf{97.5}   \\
$F^3$       & 37.2 &57.2  &   74.5           &  86.9              & \textbf{94.6}   \\
$F^4$       & 46.2 &95.0  &   92.5           &  98.0              & 96.7            \\
$F^{11}$    & 44.3 &53.4  &   43.0           &  79.6              & \textbf{90.6}   \\
$F^{13}$    & 47.6 &41.4  &   94.6           &  81.3              & 84.8            \\
$F^{14}$    & 78.4 &45.8  &   95.9           &  93.9              & \textbf{96.8}   \\
Average     & 53.5 &65.0  &   77.4           &  88.8              & \textbf{92.4}   \\
\bottomrule
\end{tabular}
\label{tab4}
\end{table}

\textbf{Case-1: fault-4,9,11}. One type and two random variations consist of these three faults which are selected. Some methods reported in the literature are employed, i.e., FDA (fisher discriminant analysis), LFDA (local fisher discriminant analysis), SLFDA (sparse local fisher discriminant analysis) and SKLFDA (sparse kernel local fisher discriminant analysis) [8]. The result of them had been illustrated in table \ref{tab1}.

\textbf{Case-2: normal and fault-2,6,7,14}. To further eliminate the influence of different methods and faulty classes on the conclusions, four faults (three types faults and one sticking fault) and one normal state are also selected to prove the effectiveness and feasibility of this proposed method. Similarly, ICA+NB (independent component analysis and na\"{\i}ve Bayes), DICA+NB (dynamic independent component analysis and na\"{\i}ve Bayes), ICA+CDMWNB (independent component analysis and class-specific distributed monitoring weighted na\"{\i}ve Bayes) and DICA+CDMWNB (dynamic independent component analysis and class-specific distributed monitoring weighted na\"{\i}ve Bayes) are employed to illustrate the feasibility of our proposed methods, and all of these methods can be detailed in literature [9].The result of them had been illustrated in table \ref{tab2}.

\textbf{Case-3: normal and fault-1,2,4,5,6,7}. To eliminate the influence of faulty types, Fault-1,2,4,5,6,7 are all step-type ones considered by literature [10] and [11] to demonstrate the good performance of their fault diagnosis methods. Similarly, SOM (self-organizing map), CCA-SOM (canonical correlation analysis and self-organizing map), and B-LSTM (Bidirectional Long Short Term Memory) literature [10] are employed to demonstrate the effectiveness and feasibility of this proposed method. Similarly, the result of them had been illustrated in table \ref{tab3}.

\textbf{Case-4: performance when classes increasing}. When comparing results in table \ref{tab1} (three classes), table \ref{tab2} (five classes), and table \ref{tab3} (seven classes), the average diagnosis accuracy decreases with classes increasing. To demonstrate this conclusion and to evaluate the impact of classes increasing, it is necessary to analyze the fault diagnosis performance when fault classes are increasing. This problem is also handled by literature [12]. The fault classes are increasing from fault-3,4,11 (3 classes) to fault-1,3,4,11,13,14 (6 classes), and the methods employed consist of FDA (fisher discriminant analysis), MRDA (maximized ratio divergence analysis), dynamic MRDA [12]. The GCN-based monitoring is implemented by ourselves, and its diagnostic results are listed in table \ref{tab4}, so do show other diagnostic results.

From these benchmark studies, four conclusions can be settled: (i) the differences in information distribution of different variables at different times can be used to describe the irregular changes to achieve process monitoring; (ii) moreover, spatial-temporal associations representation can be used to describe these irregular association changes between numerous variables; (iii) the monitoring performance of proposed method can reach to the same fault diagnosis accuracy than other methods; and (iv) the performance of the proposed method is influenced by the number of fault classes, which should be paid more attention when it is applied practically.

\subsection{Application Study}
Zinc is one of the most used non-ferrous metals, and zinc hydrometallurgy accounts for more than 85\% of zinc metal smelting. In zinc hydrometallurgy, cobalt removal from zinc ion solution is by adding the impurity removal agent (zinc elemental powder) and catalyst to reactors. Then a complex redox reaction happens with impurity metal ions. Finally, alloy or metal compound precipitation can be formed to gradually reduce the concentration of impurity ions until it tends to the range of technical indicators [13-15]. Process monitoring in this application study mainly refers to perceiving the changes during operation to illustrate the effectiveness and feasibility of this proposed method.

The cobalt removal process consists of five reactors, three heaters, two thickeners and other auxiliary equipment. Five reactors (1\#-5\#) are employed to provide the primary reaction environments, and one 6\# reactor is cascaded behind 5\# reaction as the spare reactor. Each reactor measures input flow, temperature, liquid level and electronic potential to reflect its operation state. Three heaters are employed to keep the zinc solution to the specified temperature, such as 75$^oC$. Two thickeners (1\#-2\#) are used to achieve preliminary solid-liquid separation to obtain a high-purity zinc ion solution. Forty-nine variables in the process are employed to realize its control of the long-term stable operation, as shown in table \ref{tab6}. Furthermore, this cobalt removal process mainly uses 10 control variables to change its conditions to meet the practical process demands. According to operation records of these 10 control variables, 8 conditions are defined after data preprocessing (such as local outlier factor) and employed to demonstrate the effectiveness and feasibility of this proposed method.

Similarly, this paper also employed diagnosis accuracy as the primary indicator to evaluate the performance of this proposed method. Generally, eight conditions are defined and employed to handle this demonstration by about 400 samples of the test-set and more than 700 samples of the training set. The details of parameter configuration for graph classification are also configured by the tricks obtained from the previous benchmark experiment. For the convenience of comparison and discussion, five popular methods are employed to illustrate the effectiveness and feasibility of the proposed method, such as LDA (linear discriminant analysis), KNN ($k$-nearest neighbours), LR (logistic regression classifier), DT (decision tree classifier), NB (na\"{\i}ve Bayes classifier), and graph convolutional network(GCN). And the best accuracy results of these methods are shown in table \ref{tab5}.

\begin{table}[H]
\centering
\scriptsize
\captionsetup{font={footnotesize}}
\caption{Classification performance ($Acc$) of application study (\%).}
\begin{tabular}{cccccccc}
\toprule
Class    & GCN  & LDA    &   KNN   &   LR   &   DT    &   NB     & This method     \\
\hline
$Con^1$  & 61.3  &100.0  &  100.0  & 100.0  &  100.0  &  100.0   & 98.78           \\
$Con^2$  & 53.4  &100.0  &  95.75  & 99.75  &  99.75  &  95.25   & 98.46           \\
$Con^3$  & 63.9  &100.0  &  100.0  & 100.0  &  100.0  &  100.0   & 99.87           \\
$Con^4$  & 28.4  &100.0  &  100.0  & 100.0  &  98.75  &  100.0   & 98.79           \\
$Con^5$  & 43.4  &100.0  &  99.25  & 100.0  &  100.0  &  100.0   & 99.69           \\
$Con^6$  & 43.2  &100.0  &  67.50  & 68.00  &  99.75  &  100.0   & 85.76           \\
$Con^7$  & 80.3  &100.0  &  97.00  & 100.0  &  100.0  &  100.0   & 99.05           \\
$Con^8$  & 92.4  &70.75  &  93.50  & 70.75  &  42.25  &  70.50   & 98.86           \\
Ave.  & 58.3  &96.34  &  94.13  & 92.31  &  92.56  &  95.72   & 97.41           \\
\bottomrule
\end{tabular}
\label{tab5}
\end{table}

All the results indicate that this proposed process monitoring method can extract influential associations between monitor variables to achieve considerable performance. And the considerable performances on eight conditions classification indicate three conclusions: (i) spatial-temporal associations can be used to perceive the operation changes between numerous monitoring variables; (ii) the monitoring feasibility and applicability of proposed method can be demonstrated by the similar conditions recognition accuracies of this proposed method and five popular methods; and (iii) some other methods should be combined with this proposed to improve the monitoring performance when facing some inaccessible features, which should pay more attention in future works.

\section{Further Work}
Although the effectiveness and feasibility of the proposed method are demonstrated by benchmark and application studies, some problems still need to be solved in further work. Firstly, the sensitivity of the proposed method for fault types should be studied more. Secondly, time and space complexity should be another critical issue with the limited and expensive calculation resources in practical industrial processes. Thirdly, the optimal class number each model can cover should be the unexpected event that needs to be optimized to keep the superiority of the proposed method. Fourthly, the false recognition rates should also pay more attention to, and this problem has yet to be deeply studied. Fifthly, appropriate parameters calculations need the a whole system theory including the graph convolutional neural network (whose parameters are not optimized in this paper), and other graph-based monitoring methods.

Finally, for more technical details, please refer to the article in the journal "Process Safety and Environmental Protection", and it only document our issues and relevant considerations for solutions of our interests.

\section*{Acknowledgment}
This work was supported in part by the Major Key Project of PCL(Peng Cheng Laboratory) under Grants PCL2023AS7-1, and also supported in part by the National Natural Science Foundation of China under Grants 62103207. Moreover, the details can refer to the paper published in the journal of "Process Safety and Environmental Protection".

\end{multicols}
\end{document}